\documentclass[sn-mathphys]{sn-jnl}

\usepackage{graphicx}%
\usepackage{multirow}%
\usepackage{amsmath,amssymb,amsfonts}%
\usepackage{amsthm}%
\usepackage{mathrsfs}%
\usepackage[title]{appendix}%
\usepackage{xcolor}%
\usepackage{textcomp}%
\usepackage{manyfoot}%
\usepackage{booktabs}%
\usepackage{algorithm}%
\usepackage{algorithmicx}%
\usepackage{algpseudocode}%
\usepackage{listings}%
\usepackage{comment}%
\usepackage{natbib}
\setcitestyle{aysep={}}

\theoremstyle{thmstyleone}%
\theoremstyle{thmstyletwo}%

\theoremstyle{thmstylethree}%

\begin{document}

\title[Article Title]{Text clustering applied to data augmentation in legal contexts}


\renewcommand{\thefootnote}{\arabic{footnote}}

\author*[1]{\fnm{Lucas}\sur{Freitas}}\email{lucas.freitas@stf.jus.br}

\author[2]{\fnm{Thaís} \sur{Rodrigues}}

\author[2]{\fnm{Guilherme} \sur{Rodrigues}}

\author[1]{\fnm{Pamella} \sur{Edokawa}}

\author[3]{\fnm{Ariane} \sur{Farias}}

\affil*[1]{\orgdiv{Strategic Management Office}, \orgname{Brazilian Supreme Federal Court}, \orgaddress{\street{Praça dos 3 Poderes}, \city{Brasília}, \postcode{70175-900}, \state{Federal District}, \country{Brazil}}}

\affil[2]{\orgdiv{Department of Statistics}, \orgname{Brasília University}, \orgaddress{\street{Darcy Ribeiro Campus}, \city{Brasília}, \postcode{70910-900}, \state{Federal District}, \country{Brazil}}}

\affil[3]{\orgdiv{Internal Affairs Department of Justice}, \orgname{Roraima State Court of Justice}, \orgaddress{\street{Praça Centro Cívico}, \city{Boa Vista}, \postcode{69301-380}, \state{Roraima}, \country{Brazil}}}


\abstract{Data analysis and machine learning play crucial roles in legal domains, particularly in clustering and text classification tasks. This study employed natural language processing tools to augment datasets meticulously curated by experts, improving classification flows of legal texts through machine learning. We considered the Sustainable Development Goals (SDGs) data from the United Nations 2030 Agenda as a practical case study. The employed data augmentation strategy has yielded substantial improvements in the accuracy and sensitivity metrics of classification models, with some SDGs of the 2030 Agenda showing gains exceeding 15\%. In certain instances, the example base has expanded by a remarkable factor of 5. When unclassified texts are available for certain legal labels, data augmentation strategies centered around clustering are effective options for augmenting the available knowledge base, thus reducing labor-intensive manual classification efforts.
}




\maketitle

\section{Introduction}\label{sec1}

Currently, courts around the world are adopting open data policies to promote greater transparency and, consequently, increase access to justice. This recent effort by courts to automate the publication of legal data through APIs and dashboards also brings numerous benefits to natural language processing (NLP) activities. NLP is a computer science field that focuses on enabling computers and machines to comprehend human languages and texts. With the advent of new computational techniques and large-scale data produced by legal operators, computers are now capable of grouping, classifying, generating text, and performing deep statistical analyses on legal texts \citep{bart2022, ashley2017}.

Large legal databases and automation workflows in legal domains have been operating on dedicated terminals since the mid-1970s (examples include Westlaw and QUIC/LAW) to facilitate the electronic provision of legal documents. After technological advances, databases such as LexisNexis can be used today in mobile phone applications. Through robust APIs or simple queries in the palm of one's hand, the goals of gathering and utilizing large sets of legal data remain the same, encompassing the provision of rapid searches (information retrieval) and the support for decision-making processes in both law firms and courts.

Structured data within environments equipped with robust search tools are part of what is needed, alongside powerful computers and expert teams, to incorporate machine learning into legal contexts \citep{dimatteo2022}. Machine learning is at the crossroads of computer science, mathematics, and statistics, endowing computers with the ability to learn certain tasks by discerning patterns in training datasets \citep{james2014}. When input data are organized into texts, these tasks naturally fall within the realm of natural language processing.

In this study, we address the significant potential of utilizing natural language processing for legal text classification tasks, especially within the context of the Brazilian Supreme Federal Court (STF). Specifically, we demonstrate that employing data clustering techniques and data augmentation on expert-labeled databases can enhance the performance of NLP models in classifying tasks related to the United Nations' 2030 Agenda for Sustainable Development (SDGs).

This article is organized as follows. The next section discusses the importance of internalizing the UN 2030 Agenda in courts and highlights positive aspects of the Brazilian Supreme Federal Court experience. Section \ref{sec3} discusses previous work on machine learning and text classification in legal domains. Section \ref{sec4} demonstrates the proposed clustering method to improve text classification workflows for the SDGs of the 2030 Agenda. Section \ref{sec5} is dedicated to experiments and results, while the final two sections focus on discussions and conclusions.

\section{UN 2030 Agenda}\label{sec2}

The 2030 Agenda of the United Nations is a global pact with the aim of improving the world for all peoples and nations by 2030. Established in September 2015 with the participation of 193 signatory countries, it encompasses 17 Sustainable Development Goals (SDGs) related to zero hunger, quality education, gender equality, clean water, reduced inequalities, peace, and justice, among others. These 17 SDGs also encompass 169 universal targets meticulously formulated after extensive international public consultation, with a dedicated focus on addressing the needs of the most vulnerable individuals and nations. Numerous studies have explored the interdependencies of the SDGs within the UN's 2030 Agenda, acknowledging that actions targeting one goal can inherently yield positive effects on the targets of others \citep{kroll2019, breuer2019}.

International commitments of this magnitude require active involvement from all sectors of society, as well as from every government agency within the signatory countries. In Brazil, the Supreme Federal Court has been making efforts to incorporate the 2030 Agenda into its activities since 2020, particularly through the manual classification of constitutional control legal processes and cases with recognized general significance into the SDGs agenda. The court newsletter also categorizes cases under the SDGs of the 2030 Agenda, promoting correlations between judgments and the SDGs. Such data will be used in this study for machine learning applications to support process classification.

By classifying cases according to the SDGs of the 2030 Agenda, opportunities arise to compare the actions of the Supreme Federal Court with other international constitutional courts. Furthermore, any court can use the sustainable development goals to prioritize judgments and formulate indicators. Examples of implementing the objectives outlined in the UN's 2030 Agenda within the public sector are evident in various countries such as Canada, the European Union, Germany, Italy, Switzerland, and Australia. Starting in 2021, the Brazilian Supreme Federal Court initiated the classification of some cases using a machine learning tool named RAFA 2030. This tool not only offers recommendations for the SDGs, but also incorporates a range of graphs to enhance the comprehension of the analyzed text and its subsequent handling.


\section{Background}\label{sec3}

For many years, legal researchers have employed manual and exhaustive methods to bring new interpretations to legislation and tackle practical issues in courts and law firms. Cases and processes were organized in large volumes, each containing countless pages of typed or handwritten content \citep{epstein2010}. This kind of research, recognized as doctrinal research, has objectives encompassing the meticulous elucidation of regulations relevant to a specific legal domain, coupled with the clarification of the legal framework's significance via analyses of laws and case law. Before the advent of computers, these researchers could only depend on reliable legal sources within the area under scrutiny, refraining from extrapolating relationships or making predictions about the future state of law and case law \citep{chui2017}.

Statistical and quantitative methods faced challenges when applied to doctrinal research when processes were physical or paper-based. Extracting information from physical documents was very costly, making extensive research and consultations impossible within constrained timeframes  \citep{epstein2014}. Currently, with courts' efforts to disclose legal process data online, it is unfeasible to comprehensively read and organize all available information on a specific legal subject, particularly in scenarios where international decisions and case law come into play. With the increase in the production of large-scale legal databases and the progress of computational and mathematical methods to handle large datasets, the fusion of statistical techniques for empirical research and doctrinal legal research is now entirely viable. This integration not only unveils novel patterns, but also broadens the scope of applied legal analysis \citep{derlen2017, goanta2017}. 

Even with the newfound capability to process legal data aided by computational tools, certain quantitative studies still rely on databases painstakingly curated by legal experts. This approach typically involves investigating the relationships between decisions and social dimensions such as politics, race, and gender \citep{rachlinski2017, frankenreiter2016}. In the United States (US), a nation renowned for its robust tradition of incorporating quantitative analyses into law, a noteworthy study by Katz et al. emerges \citep{katz2017}. This study combined 240,000 votes and 28,000 cases meticulously annotated by human experts, all accessible in the Supreme Court Database (SCDB) \citep{spaeth2022}. The objective was to forecast the behavior of the US Supreme Court in its judicial rulings.

Currently, quantitative methods are employed in legal research worldwide. Examples can be found in various countries, including Belgium \citep{de2017}, France \citep{sulea2017}, the Netherlands \citep{vols2017}, South Africa \citep{lefakane2022}, Argentina \citep{prometea2019}, Israel \citep{doron2015} and India \citep{mandal2021}. In Brazil, the contributions of \cite{deoliveira2023, hartmann2019, feijo2023, vianna2023, jacob2022, sabo2022, nunes2022} stand out. In common, these Brazilian works use human-labeled databases for legal classification, clustering, summarization, and topic modeling activities in first-instance (state) and second-instance or special (federal) courts.

Among the quantitative methods most used in the legal context, approaches based on data science and machine learning stand out \citep{surden2014}. These methods utilize learning strategies to train algorithms in repetitive tasks, such as clustering and legal text classification. When algorithms are provided with a substantial number of training examples, meaning labeled or classified legal processes and texts, supervised learning becomes the operative framework. In practice, algorithms of this nature simulate expert decisions automatically with minimal or no human intervention. 

However, when labeled processes are scarce or there is a significant label imbalance between the judicial processes under study, supervised learning strategies cannot be used directly for classification or clustering tasks \citep{geron2022}. Therefore, as in the present study, the need for new labels (classifications) arises for courts and law firms. Here the focus lies on classifying cases into SDGs of the 2030 Agenda of the UN. Another interesting example relates to the COVID-19 pandemic, during which legal cases related to the topic were prioritized in the Brazilian judicial system. One potential solution to these problems involves unsupervised learning methods or even combinations of supervised and unsupervised methods.

Unsupervised learning methods are naturally applied to tasks relevant to those carried out in courts and law firms, such as clustering similar cases for subsequent batch analysis. Within the Brazilian judiciary, notable tools include Athos from the Superior Court of Justice \citep{atos2022}, Berna from the Goiás State Court \citep{berna2020}, and vitorIA from the Supreme Federal Court. These initiatives employ unsupervised learning to create groups of similar cases based on specific document texts or metadata, enabling bulk handling of similar legal matters.

Legal case clustering initiatives encompass the creation of contextual frameworks, extending their focus beyond mining specific arguments in legal texts or predicting decisions, as evidenced in some of the referenced studies. In data-rich environments, these initiatives assume a critical role by supporting a spectrum of legal activities, including the training of new legal professionals, the management of case law in court systems, and the curation of repositories housing general legal arguments within law firms. In simpler terms, these initiatives facilitate administrative tasks like categorization and, most importantly, decision-making by grouping analogous cases. In this study, a novel approach is proposed using machine learning-based text clustering techniques and data enhancement strategies to expand the training dataset for case law classification related to the SDGs of the 2030 Agenda.

Data augmentation comprises a set of techniques primarily aimed at increasing or enhancing the diversity of training examples in machine learning and deep learning workflows, all while refraining from explicitly adding new observations to the dataset through collection \citep{feng2021} or artificial simulations (synthetic data). Within the legal context, this approach is valuable for expanding the number of examples of unusual or past occurrences. For instance, consider the Brazilian Supreme Federal Court, where issues of general repercussion tend to surface in waves of litigation, prompting analogous cases to be expeditiously addressed in lower courts often circumvents higher instances. Furthermore, data augmentation strategies can be employed to increase the number of examples of historical cases of general repercussion, thereby improving the performance of machine learning models in their identification.

To contribute to the current state of the art in data augmentation in legal domains, this work proposes a label propagation pipeline based on clustering of unlabeled and expert-labeled data, with a special focus on controlling the efficiency of augmented data through adjustments of model parameters.

\section{Methodology}\label{sec4}

Legal information is processed in various ways by legal practitioners, such as courts and law firms. However, there is a prominent trend towards integrating machine learning and artificial intelligence into legal domains, whether it is in the processing of texts for summarization, classification, grouping, or in the automation of numerous manual and repetitive activities intrinsic to legal operations.

The purpose of this study is to introduce a methodology to augment databases related to the classification of cases into the Sustainable Development Goals (SDG) of the 2030 Agenda, using larger databases of originally unlabeled texts. This enhancement has the potential to suggest corrections to knowledge bases or to improve the performance of machine learning models used in classification tasks, as explored in this research.

\begin{figure}[h]
	\begin{center}
		\includegraphics[height=8.5cm,width=13.3cm]{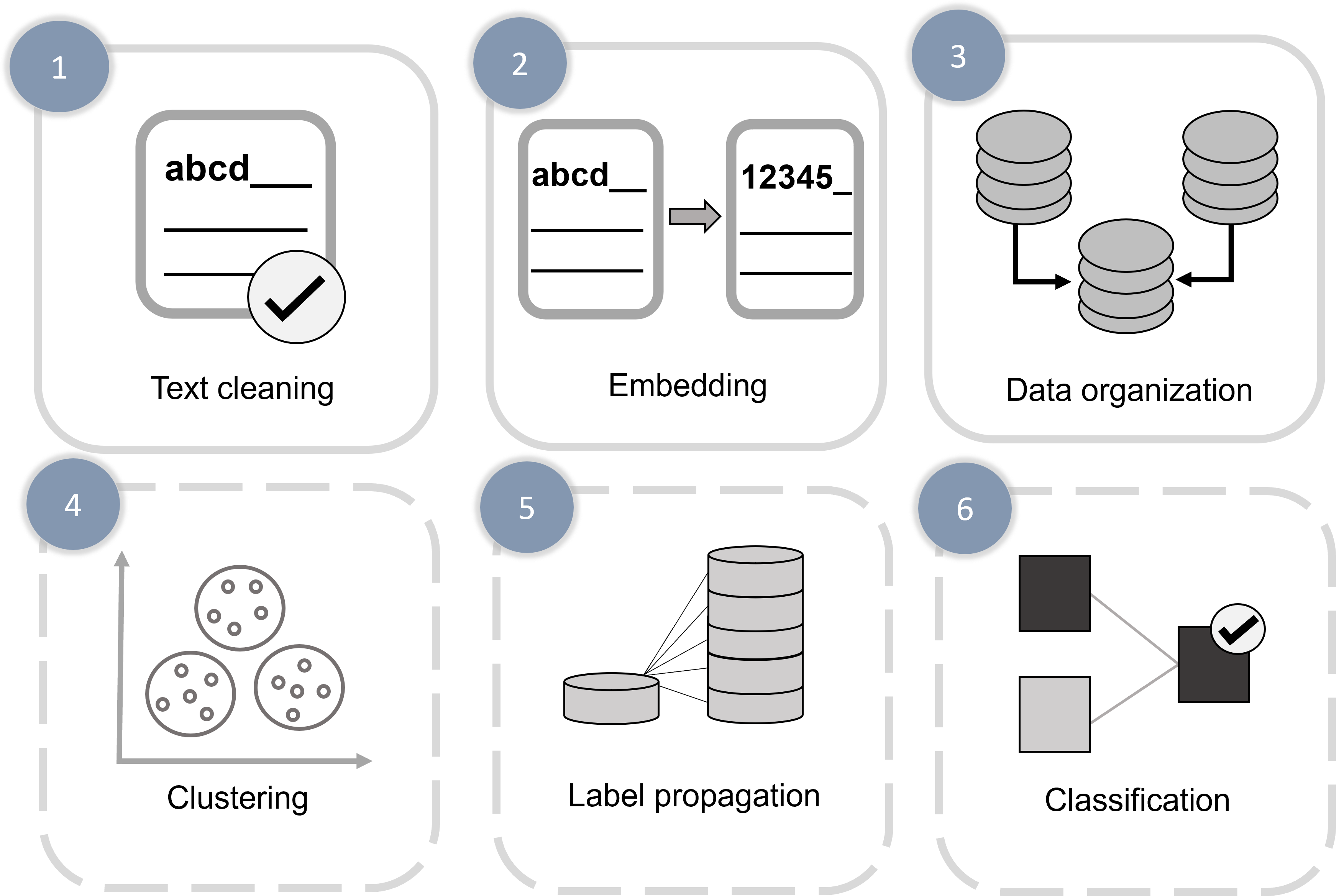}
		\vspace{0.45cm}
		\caption{Basic flowchart. Text cleaning, embedding and data organization are performed for all SDGs in batch (solid line steps), while clustering, label propagation and classification are performed individually for each of the SDGs (dashed line steps).}\label{fig:Fig1}
	\end{center}
\end{figure}

Classification activities involve supervised learning routines aimed at predicting a legal category linked to a new textual entity.  In contrast, clustering activities involve unsupervised learning routines designed to group similar text processes. In legal contexts, classification and clustering tasks can overlap, as similar processes may share metadata and labels. Processes with similar labels (procedural class, main subject, etc.) can form a group and collectively follow the same flow within a court or law firm. In this study, the integration between these two activities occurs, as the proposed clustering strategy is applied to databases that collect labeled and unlabeled processes related to the SDGs of the 2030 agenda.  This approach highlights how some initially unlabeled entities can automatically receive artificial (synthetic) labels, without the need for manual classifications. The proposed method, which encompasses data augmentation and classification, is shown in Figure \ref{fig:Fig1}.

The representation in Figure \ref{fig:Fig1} is presented at a high level, providing a general overview of the workflow proposed here. Labeled (associated - yes or no - with the SDGs of the 2030 Agenda) and unlabeled datasets are brought together into a single entity. Text cleaning, embedding, and data organization are steps that are carried out for all SDGs in batches (solid line steps in Figure \ref{fig:Fig1}). Before the clustering step, a simple upsampling process is performed to reinforce the label signals with few entries. Then, the clustering, label propagation and classification steps are performed individually for each of the SDGs (dashed line steps in Figure \ref{fig:Fig1}). The clustering mechanism is employed to create clusters of processes with similar textual content within this new framework. Disregarding the cluster boundaries, processes initially lacking labels are assigned artificial labels based on the presence or absence of labeled processes in their proximity (label propagation), increasing the number of positive examples for specific labels. The final step involves classifying the texts using neural networks trained with the augmented data.

Within the proposed workflow, new methods for preprocessing, embedding, clustering, and classification can be incorporated. This study employed simple methods to establish a baseline and present the approach in a clear way. The following subsections details the main steps of the proposed workflow.

\subsection{Embedding}\label{subsec1}

Embedding involves transforming text into numerical vectors, which are entities easily interpretable and processable by computers. Vectors are essentially a collection of objects, while vector spaces are collections of vectors in some dimension. Representing objects in vector spaces means embedding them in environments equipped with distance metrics, facilitating comparisons between these objects. Representing texts as vectors of numbers offers a means to compare texts using algorithms, models, or computational methods.

The simplest methods for representing texts in a collection (corpus) are based on word frequencies. TF-IDF method \citep{salton1988} is a prominent example of this approach. TF-IDF involves converting each text in a corpus into a vector of the size of the dictionary (total number of words) associated with the corpus. The entries of these vectors account for the frequency of each word both within the specific text and across the entire corpus. When a particular word from the dictionary does not appear in a text, the corresponding vector position is filled with a zero value. This kind of approach generates sparse and high-dimensional vectors, since not all words will appear in all texts of the collection, and small corpus word dictionaries can easily exceed one million words.

Yet another challenge posed by frequency-based embedding techniques is that texts with similar meanings may end up having distinct vectors. This discrepancy arises from variations in word frequency, arrangement, and the usage of synonyms. Since these frequency-based embeddings neglect contextual nuances in text representations, they encounter difficulties in accurately capturing writing styles, regional idioms, and tend to perform poorly when applied to an extensively diverse corpus, i.e., those with a wide array of topics \citep{jurafsky2018}. The word2vec family, with particular emphasis on the doc2vec method, addresses these problems.

Word2vec approaches use two-layer neural networks to artificially represent words hidden according to their neighboring words (continuous bag of words – CBOW), or to predict the neighborhood based on central words (skip-gram) using the neural network's own weights \citep{mikolov2013}. The doc2vec method extends word representations from word2vec to texts without constraints on size. The theoretical novelty of doc2vec \citep{le2014} lies in training the textual representation vector, via weight updates, in conjunction with the word representation vectors that collectively form the text itself. In this work, the texts will be vectorized using the doc2vec method.

\subsection{Clustering and label propagation}\label{subsec2}

Text grouping typically involves making comparisons among vectors derived from the embedding phase. Specifically, in legal contexts, text clustering activities serve the purpose of streamlining repetitive and manual tasks, as similar cases can be processed more efficiently in batches. However, in this study, the clustering step aims to extend labels to originally unclassified texts for the 2030 Agenda SDGs. 

The approach proposed here is applicable in scenarios where there is a large collection of unclassified texts available, but labeled texts are scarce and the classification task demands substantial resources. In the case of the study, processes began to be labeled with SDG tags from the 2030 Agenda only in 2020, even though there are many usable texts (cases) predating the 2030 Agenda itself. The goal of assigning labels to new texts is to reinforce signals, especially for less common labels, and enhance the performance of classification algorithms. For instance, SDG 16, focusing on peace, justice, and strong institutions, exhibits significantly more cases than others. This is explained by the stronger correlation of SDG 16 - Peace, Justice and Strong Institutions - with the nature of constitutional courts like the Brazilian Supreme Federal Court. The original label distribution can be seen in Table \ref{tab1}.

\begin{table}[h]
	\caption{Label distribution for manually labeled processes.}\label{tab1}%
	\begin{tabular}{@{}lcc@{}}
		\toprule
	SDG	& Labels 0 & Labels 1 \\ \midrule
		SDG 3  & 1635        & 370         \\ 
		SDG 4  & 1877        & 128         \\ 
		SDG 8  & 1559        & 446         \\ 
		SDG 9  & 1937        & 68          \\ 
		SDG 10 & 1635        & 370         \\ 
		SDG 11 & 1914        & 91          \\ 
		SDG 15 & 1909        & 96          \\ 
		SDG 16 & 763         & 1242        \\ 
		SDG 17 & 1787        & 218         \\ 
\botrule  
\end{tabular}
\end{table}

\begin{figure}[h!] 
	\centering
	\includegraphics[width=0.56\linewidth]{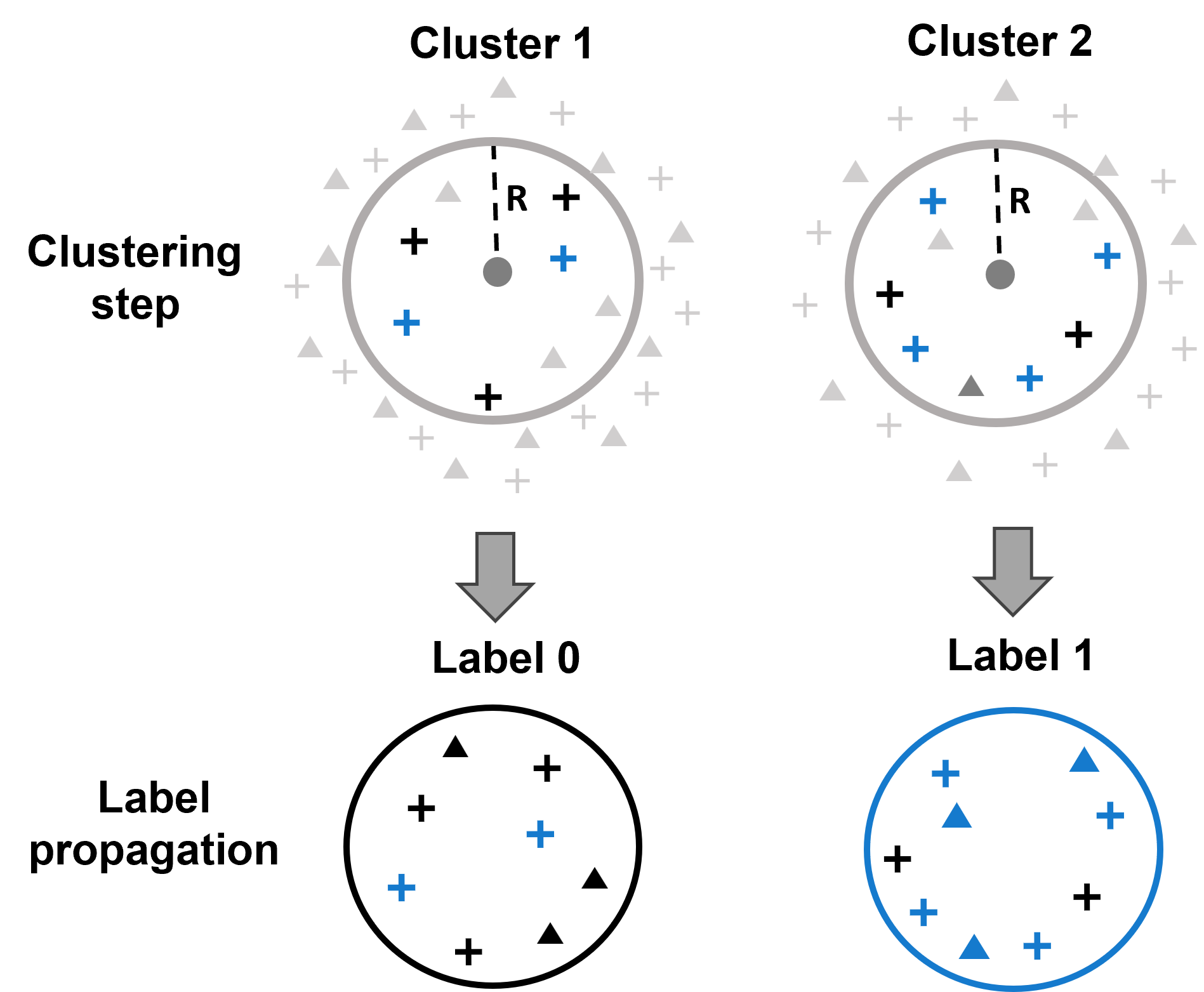}	
	\caption{Clustering strategy to data augmentation. For exemplification, data was divided into 2 clusters and unlabeled legal processes (gray triangle) in the radius R were selected. Crosses illustrate originally labeled text (black cross denotes label 0 and blue cross denotes label 1). Label propagation is performed according to the proportion of labeled processes in each selected region. Synthetic label 0 was assigned to the first cluster (black triangles) and synthetic label 1 was assigned to the second cluster (blue triangles).}\label{fig:Fig2}
\end{figure}

Table \ref{tab1} shows that there is only a limited number of entries available for training an artificial neural network to perform the SDG classification. In this context, it is important to consider data augmentation strategies. In this study, the proposed approach involves merging the original labeled texts with an unlabeled dataset, followed by the use of a clustering algorithm to assign synthetic labels within well-established groups. A portion of the labeled dataset (60\%) is dedicated to the training or adjustment stage of clusters using k-means, along with the entire unlabeled dataset. Altogether, the number of cases used for clustering in this training stage varies around 44000 entries. 20\% of the labeled data set is used for the validation of the clustering parameters, and the remaining 20\% is used to test the clustering algorithm. Figure \ref{fig:Fig2} illustrates the clustering strategy.

The diagram showcases both originally labeled texts (depicted as crosses) and processes yet to be assessed for the designated SDG (represented as triangles). Both data sets are grouped using the k-means \citep{macqueen1967, tan2018} clustering algorithm based on their text embeddings. K-means is a straightforward algorithm that defines a prototype using a centroid, typically the mean of a group of points. It is commonly applied to data objects in a continuous n-dimensional space. Initially, we select K initial centroids, with K being a user-specified parameter representing the desired number of clusters. Each point is assigned to the nearest centroid, forming clusters. Then the centroid of each cluster is updated based on the points within it. This assignment and update process continues until no point changes clusters or until the centroids stabilize. For simplicity, the figure presents only two groups, although the number of clusters can be adjusted as a model parameter. 

For the label propagation step, we propose to assign synthetic labels only to processes within the vicinity of the cluster center, and by avoiding the cluster edges, greater similarity among processes in the group is achieved. Within this proximity (gray circle with radius R), we have processes associated with the specified SDG (blue crosses denoting label 1), processes not associated with the SDG (black crosses denoting label 0), and processes that have not been evaluated yet (gray triangles). 

Figure \ref{fig:Fig2} also shows the assignment of labels (colors) to certain unlabeled processes. Unclassified processes (triangles) located within the radius of the circle are assigned synthetic labels on the basis of the proportion of original labels in their vicinity. The threshold proportion also serves as a model parameter that requires calibration, as 50\% may not always be the optimal choice. In Figure \ref{fig:Fig2}, for explanatory purposes, we considered the criterion of at least 50\% positive labels (processes related to the specific SDG) within the vicinity to assign label 1 to nearby unclassified processes. This means that unlabeled processes within the vicinity of cluster 1 received label 0 (gray triangles turned black), while those within the vicinity of cluster 2 received label 1 (gray triangles turned blue).

These synthetic labels increase the number of examples, completing the text-based data augmentation strategy for initially uncategorized data. Given unlabeled document collections, the approach proposed in this study emerges as an alternative to well-known data augmentation methods, such as upsample and downsample techniques, as well as the previously mentioned easy data augmentation and back translation approaches \citep{feng2021}.

\subsection{Text classification}\label{subsec3}

In machine learning, text classification involves automatically labeling texts using algorithms based on predefined categories. The tasks of classifying and grouping texts share similarities, since it is possible to assign cluster labels to objects within a group. Classification activities generally rely on supervision, depending on prior categorizations. On the other hand, clustering activity involves grouping texts with similar vector representations (embeddings), constituting an unsupervised task.

In a legal context, the classification of legal documents is a fundamental activity within courts and law firms. During the initial stages of legal proceedings, legal professionals perform classification activities to set the process in motion. As cases progress, these classification tasks come into play to assign documents and actions to their respective collaborators and departments. Ultimately, cases are categorized to establish organized historical archives and to streamline searches for legal precedents and institutional records. The extraction of metadata derived from classifications and procedural classes (categories)\footnotemark[1], for instance, is utilized to compare the performance of courts, as exemplified by the UN's 2030 Agenda Sustainable Development Goals (SDGs).

\footnotetext[1]{The Supreme Federal Court of Brazil (STF) handles various procedural categories, including appellate and original categories. Extraordinary appeals (RE) and extraordinary appeal motions (ARE) constitute the largest workload. REs originate from cases in lower courts and focus on constitutional issues. Before the introduction of the general repercussion judgment system, STF decisions in these cases didn't bind lower courts. Original categories encompass constitutional law review and individual constitutional guarantees procedures.}

There are numerous methods for text classification based on machine learning and deep learning \citep{eisenstein2019}. Algorithms such as Naive Bayes, Support Vector Machines (SVM) \citep{hvitfeldt2021}, Extreme Gradient Boosting (XGBoost) \citep{chen2016}, and CatBoost are widely used for text labeling. CatBoost, in particular, represents a significant recent improvement as it combines texts and metadata without relying on voting schemes \citep{prokhorenkova2019}.

Deep neural networks are also widely used for text classification, with architectures such as LSTM and CNN, adapted from image classification, emerging as the most preferred choices \citep{kamath2019}.
Since the goal of the data augmentation strategy proposed in this study is to enhance classification model performance, augmented datasets containing synthetic labels are used to train classification algorithms. The idea is to assess whether an increased number of positive label examples impacts textual classification models, particularly for labels with limited original records. The next chapter is dedicated to the conducted experiment, presenting associated results and model parameters choices.

\section{Experiments and Results}\label{sec5}

The series of experiments carried out here encompasses studies on the optimal parameters for clustering, a task associated with data enhancement activity. Additionally, these experiments explore the fitting of text classification algorithms to determine whether the augmented datasets, generated using the proposed method, improve the performance of the classification task itself. The following section details the dataset.

\subsection{Data}\label{subsec5-1}

The current study uses two datasets: one with labels corresponding to the Sustainable Development Goals (SDGs) of the 2030 Agenda, and another without such labels. The labeled dataset comprises approximately 2000 texts, including judgments and initial petitions from cases within the Supreme Federal Court (STF). Judgments are records of collective court rulings, whereas initial petitions, as the name implies, encompass the initial requests submitted by parties when they initiate legal proceedings in the judicial system.

The original classification (label used in supervised learning procedures) is conducted by the court's own staff through the examination of two documents: judgments and petitions. This classification is carried out retrospectively, meaning it occurs after the legal proceedings have begun and decisions have been reached. Petitions and judgments play a vital role in the case classification process, as they contain initial claims and rulings, along with ministers' opinions on the specific case at hand. Since classification is based on an analysis of these two documents, there is no risk of a single case receiving differing labels in its petition and judgment. The allocation of SDG labels in the labeled dataset was previously outlined in Table \ref{tab1}.

Label 1 indicates that the case examined is related to the specific SDG, while label 0 indicates that the case has been reviewed and does not have any association with the SDG. Considering that each case can carry multiple labels, the sum of each row in Table \ref{tab1} represents the total count of cases in the labeled data set. The limited occurrences of label 1 and the imbalance between the quantities of label 1 and label 0 justify the data augmentation strategy proposed in this article.

The unlabeled dataset comprises more than 40,000 Supreme Federal Court judgments between 2015 and 2018. These are cases that have not been evaluated in terms of the SDGs, as such labels began being used in 2020 and are not mandatory for initiating judicial proceedings. These texts will be employed in this study to enrich the initially labeled dataset, introducing new contexts/words and addressing the original data imbalance.

All documents can be accessed through the official website of the court \citep{stf}, provided that they are not subject to judicial confidentiality and their disclosure does not violate the Brazilian General Data Protection Law (LGPD) \citep{soares2022}. The preference for judgments in the unlabeled dataset is justified by the fact that such documents represent the final legal decisions and are more conveniently retrievable as PDF texts, due to their standardized format established by the Supreme Federal Court.

\subsection{Parameter selection}\label{subsec5-2}

The parameters required for the clustering step are the number of clusters, centroid distance radius, and the threshold proportion for label propagation. While some ranges of values were established for each parameter based on the dataset at hand, practical applications may require broader intervals at the start of calibration, before refining the search space. Here, the models were fitted considering the parameter variations within the designated grid. Furthermore, the selection of the best model took into account optimizing accuracy and sensitivity metrics.

In data augmentation strategies focused on non-hierarchical clustering algorithms, the choice of the number of clusters plays a pivotal role in shaping the distribution of original labels within the generated groups. A limited number of clusters could lead to a dense concentration of original labels in each group, potentially causing overly generalized or less precise label propagation. Conversely, employing a larger number of clusters may result in fewer original labels per group, potentially leading to the generation of inaccurate synthetic labels. In this study, we fine-tuned the k-means algorithm across different settings, specifically exploring 5, 10, 25, 50, and 100 clusters to determine the optimal configuration.

In data augmentation strategies focused on non-hierarchical clustering algorithms, the choice of the number of clusters can have a significant impact on the distribution of original labels within the generated groups. A small number of clusters can result in unbalanced concentrations of original labels per group, potentially leading to more generalized or less precise label propagation. Conversely, with a large number of clusters, there might not be a significant number of original labels per group, resulting in inaccurate synthetic labels. In this study, the k-means algorithms were tuned for 5, 10, 25, 50, and 100 clusters.

The centroid distance radius determines how far from the cluster edges candidate processes for receiving synthetic labels are positioned. Larger radii indicate that more processes will receive synthetic labels during the propagation step, while the opposite occurs with smaller radii. The idea behind avoiding the cluster edges is to exclude texts that may have relevant associations with more than one cluster, implying higher uncertainty in their classification. The selected radii were 5\%, 10\%, 25\%, and 100\%. A 5\% distance implies that only the closest 5\% of processes to the centroid were retained, whereas a 100\% radius means that all processes within the cluster were retained for the label propagation step.

The classification threshold proportion plays an important role during the label propagation step. It defines the decision-making rule associated with label propagation. When examining processes within the cluster radius, the proportion of occurrences of label 1 among the originally labeled processes is evaluated. If this proportion is high, surpassing the classification threshold proportion, then label 1 is extended to all unlabeled processes within the same radius. If the ratio of labeled processes with label 1 falls below the threshold, then synthetic label 0 is assigned to the unlabeled processes within the radius. The thresholds under evaluation here were 50\%, 60\%, and 70\%. The rule tied to the threshold 50\% is straightforward: if more than 50\% of processes within the radius carry label 1 (indicating the presence of SDGs), the same label is applied to processes initially unlabeled within the radius. The same rule applies to the thresholds of 60\% and 70\%, which are designed to mitigate false positive labels.

For parameter selection, a validation dataset containing 20\% of the processes of the labeled dataset was used. This means that these processes labels were masked during the clustering step, and the labels assigned during modeling were compared to the original masked labels. In the experiment, all possible combinations of cluster number, centroid distance radius, and propagation threshold proportion were evaluated. Table \ref{tab2} presents the optimal parameter configuration for each SDG.

\begin{table}[h]
	\caption{Parameter choices and evaluation metrics during the validation phase.}\label{tab2}%
	 \begin{tabular*}{\textwidth}{@{\extracolsep\fill}lccccc@{}}
	\toprule
SDG		& Clusters & Distance(\%) & Threshold(\%) & Accuracy & Sensitivity \\ \midrule
		SDG 3  & 25       & 10      & 60   & 0.78     & 0.78          \\ 
		SDG 4  & 25       & 10      & 60   & 0.79     & 0.75          \\
		SDG 8  & 25       & 10      & 70   & 0.78     & 0.74          \\ 
		SDG 9  & 25       & 10      & 70   & 0.66     & 0.67          \\ 
		SDG 10 & 25       & 10      & 70   & 0.81     & 0.82          \\ 
		SDG 11 & 25       & 10      & 70   & 0.70     & 0.70          \\ 
		SDG 15 & 25       & 10      & 70   & 0.81     & 0.78          \\ 
		SDG 16 & 25       & 25      & 60   & 0.79     & 0.82          \\ 
		SDG 17 & 50       & 25      & 60   & 0.65     & 0.62          \\ 
	\botrule
 \end{tabular*}
\end{table}

After parameter selection, 20\% of the remaining labeled data set is used for the testing phase. This approach seeks to assess the performance of clustering on a distinct dataset after identifying the optimal parameters for each label. The proposed clustering strategy for label propagation has the advantage of controlling the classification efficiency, preventing incorrect label propagation in poor clusters. Further details on annotated data control in machine learning workflows can be found in \citep{ant2023}. Table \ref{tab3} presents the accuracy and sensitivity metrics for each SDG during the testing phase. \\

\begin{table}[h]
	\caption{Evaluation metrics during the testing phase.}\label{tab3}%
	\begin{tabular}{@{}lcc@{}}
		\toprule
        SDG	& Accuracy & Sensitivity \\ \midrule
		SDG 3  & 0.77     & 0.77          \\ 
		SDG 4  & 0.75     & 0.73          \\ 
		SDG 8  & 0.75     & 0.72          \\ 
		SDG 9  & 0.73     & 0.78          \\ 
		SDG 10 & 0.81     & 0.76          \\ 
		SDG 11 & 0.72     & 0.73          \\ 
		SDG 15 & 0.74     & 0.71          \\ 
		SDG 16 & 0.79     & 0.80          \\ 
		SDG 17 & 0.65     & 0.66          \\ 
\botrule
 \end{tabular}
\end{table}
\vspace{-0.32cm}

It is possible to observe that SDGs with a broader scope often exhibit poorer performance, as exemplified by SDG 17 - Partnerships for the Goals. Additionally, the model performance remains consistent between the test and validation data sets. Following the upsampling procedures and data augmentation through label propagation via clustering, new label distribution can be seen in Table \ref{tab4}. \newpage

\begin{table}[h]
	\caption{Label distribution after data augmentation.}\label{tab4}%
	\begin{tabular}{@{}lccc@{}}
		\toprule
SDG		& Labels 0 & Labels 1 & Total  \\ \midrule
		SDG 3  & 3590        & 590         & 4180   \\ 
		SDG 4  & 3908        & 509         & 4417   \\ 
		SDG 8  & 3453        & 654         & 4107   \\ 
		SDG 9  & 3964        & 548         & 4512   \\ 
		SDG 10 & 3604        & 642         & 4246   \\ 
		SDG 11 & 3953        & 535         & 4488   \\ 
		SDG 15 & 3934        & 529         & 4463   \\ 
		SDG 16 & 3957        & 6438        & 10395  \\ 
		SDG 17 & 3692        & 663         & 4355   \\ 
\botrule  
\end{tabular}
\end{table}

From Table \ref{tab4}, we note that all SDGs experienced a significant increase in the number of positive examples, i.e., those with label 1. For instance, SDGs 9 and 11 now have over 500 instances with label 1, compared to the original count of less than 100, as presented in Table 1. This augmentation holds particular significance as it effectively doubles the overall size of the datasets. With the augmented datasets in place, the classification phase can be addressed, which is the subject of the next section.

\subsection{Text classification algorithm}\label{subsec5-3}

LSTM architecture networks are widely used in natural language processing due to their ability to better manage long-term dependencies. In essence, they can effectively capture relationships among distant segments within a given text. The memory mechanism of these networks combines three different types of gates – namely, input gates, output gates, and forget gates – to determine what information to remember and what to forget in each layer of the network.

LSTM networks were chosen for the classification stage, as they are already employed in the classification of legal texts within the context of the 2030 Agenda's Sustainable Development Goals in the Brazilian judicial system.  In this study, LSTM networks were fitted for both the original and augmented datasets using the PyTorch framework \citep{pytorch2019}. The syntax of the PyTorch library is distinctive, and its embedding mechanism for adjusting LSTM networks is based on one-hot encoding (Bag of Words) and weights initialized arbitrarily. Embeddings obtained through methods like word2vec or doc2vec are not recommended.

During the forward pass, LSTM layers of the networks receive the outputs from the embedding stage and undergo consecutive dimension reductions. These reductions take into account weights for distant term distances, dropout layers, and activation layers. After completing the network's execution, the model returns probabilities of belonging to the classes of the proposed problem. In the context of this study, class 0 indicates that the text has no association with the evaluated Sustainable Development Goal (SDG) and class 1 otherwise. By default, a threshold of 0.5 is considered, where probabilities exceeding 0.5 indicate label 1.

Networks were trained using 1000 epochs and Adam optimizer with a learning rate of 0.001. The chosen loss function was Binary Cross-Entropy (BCELoss). Additionally, two dropout layers were integrated into the architecture, with dropout probabilities of 0.8 and 0.6. Further details regarding the parameters and architecture of the LSTM networks are available in \citep{morris2019}, while general graphical explanations about deep neural networks can be found in \citep{yoshua2016} and \citep{glassner2021}.

The computational cost of training LSTM networks is significantly high, particularly for augmented datasets. Consequently, instances of Google Colab Pro+ were used. Google Colab is a cloud-based development environment with essential Python packages pre-installed. On average, computational times were approximately 59 minutes, benefiting from hardware acceleration through TPUs V2 and 52GB of RAM. In the case of the original dataset, this time was under 10 minutes per SDG.

The main objective of this study is to assess whether augmented datasets, following the proposed method, enhance the performance of neural networks originally trained on datasets without artificial labels. Just as in earlier stages of the proposed workflow, the Sustainable Development Goals (SDGs) under investigation were addressed individually. This involves training a distinct LSTM neural network for each SDG.

\subsection{Results}\label{subsec5-4}

The classification tests were carried out using LSTM networks trained on both the original dataset and the augmented dataset (comprising original data as well as synthetically generated data from the clustering stage). Network performance was evaluated using a bootstrap strategy \citep{james2014} with 10,000 iterations. In this process, 80\% of the dataset was resampled with replacement for training purposes, while the remaining data not used in the training samples was considered for testing. Table \ref{tab5} presents the average accuracy and sensitivity metrics for both the original and augmented datasets. \\

\begin{table}[h]
    \caption{Average performance of LSTM neural networks trained on the original and augmented datasets.}\label{tab5}
    \begin{tabular*}{\textwidth}{@{\extracolsep\fill}lcccc}
        \toprule
        & \multicolumn{2}{c}{Original Dataset} & \multicolumn{2}{c}{Augmented Dataset} \\
        \cmidrule(lr){2-3} \cmidrule(lr){4-5}
        SDG & Accuracy & Sensitivity & Accuracy & Sensitivity \\
        \midrule
        SDG 3 & 0.83 & 0.80 & 0.89 & 0.82 \\
        SDG 4 & 0.79 & 0.83 & 0.84 & 0.81 \\
        SDG 8 & 0.86 & 0.81 & 0.87 & 0.83 \\
        SDG 9 & 0.81 & 0.79 & 0.89 & 0.87 \\
        SDG 10 & 0.83 & 0.79 & 0.85 & 0.79 \\
        SDG 11 & 0.78 & 0.75 & 0.82 & 0.81 \\
        SDG 15 & 0.72 & 0.72 & 0.83 & 0.83 \\
        SDG 16 & 0.87 & 0.82 & 0.91 & 0.85 \\
        SDG 17 & 0.73 & 0.75 & 0.74 & 0.76 \\
        \botrule
    \end{tabular*}
\end{table}

From Table \ref{tab5}, we note that all SDGs exhibited greater accuracy when using augmented data, with a noteworthy mention of SDG 15 - Life on Land, which shows a 17\% increase in this measure. The average accuracies and sensitivities exceed 80\% for all SDGs, except for SDG 17. The sensitivity metric is of particular importance, as one of the objectives of this study is to amplify the volume of texts associated with SDGs that inherently have limited examples. 

T tests \citep{student1908} used to compare the bootstrap samples show a significant improvement, at 5\% significance level, in both the mean accuracy and sensitivity measures for the augmented datasets compared to the original datasets in almost all SDGs. An exception is noted for SDGs 4 and 10, where average sensitivities slightly decreased in the augmented data set. Consequently, the methodology proposed for dataset augmentation via clustering has proven effective, leading to a substantial improvement in the classification neural network's performance for nearly all SDGs.

\begin{figure}[h]
	\begin{center}
		\includegraphics[height=5cm,width=13cm]{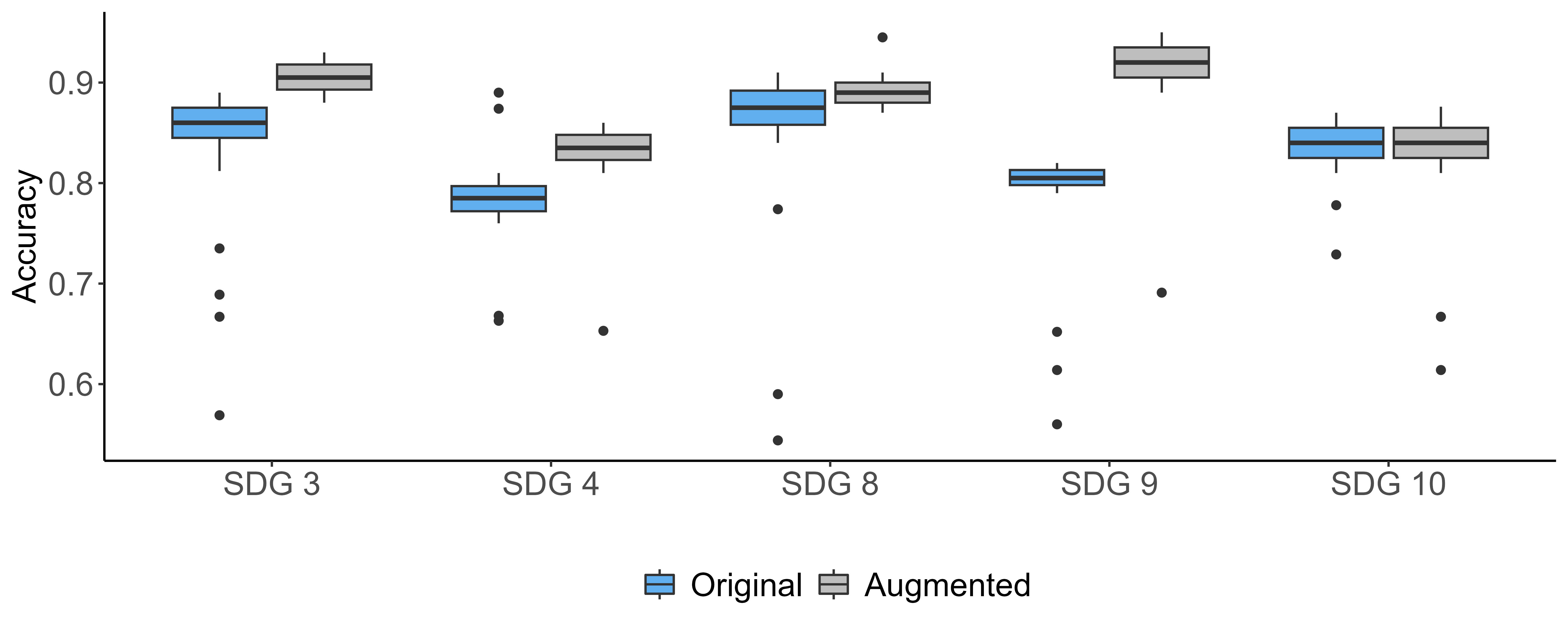}
		\caption{Accuracy for bootstrap samples from original and augmented datasets for SDGs 3, 4, 8, 9 and 10.}\label{fig:Fig3}
	\end{center}
\end{figure}

\begin{figure}[h]
	\begin{center}
		\includegraphics[height=5cm,width=13cm]{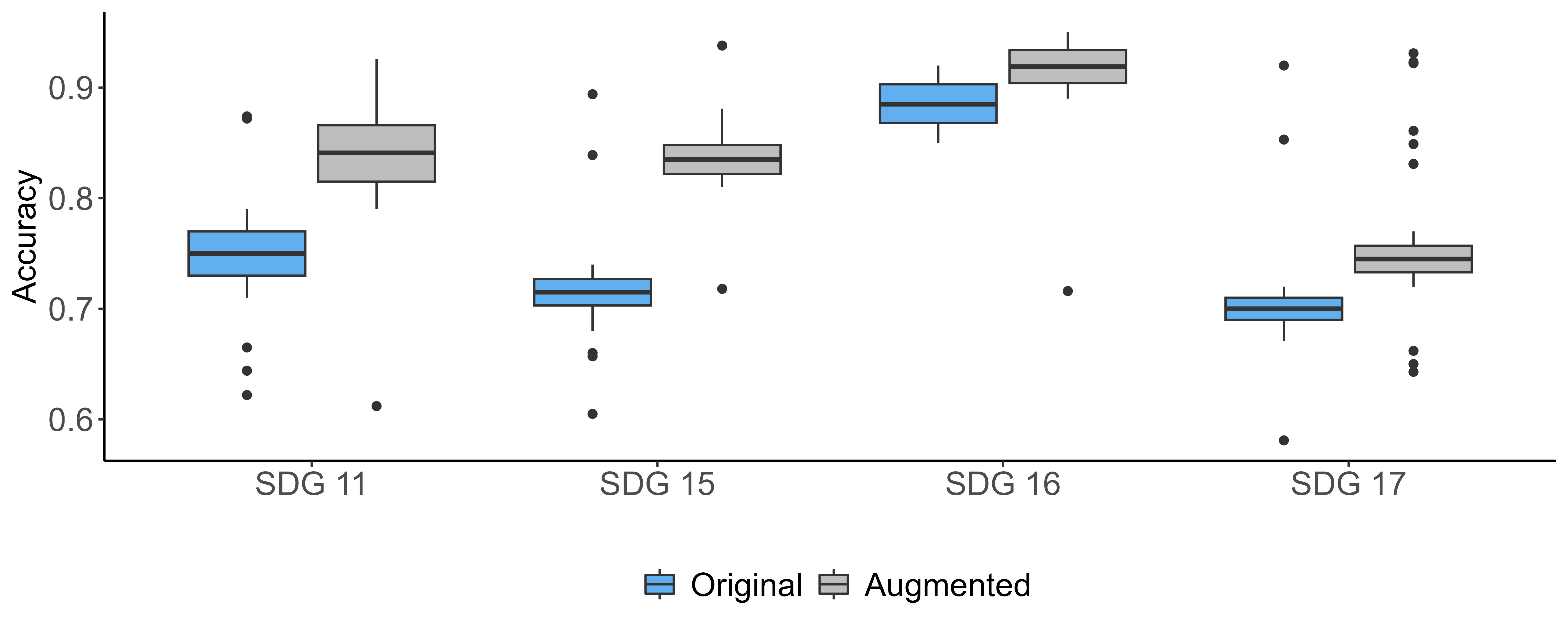}
		\caption{Accuracy for bootstrap samples from original and augmented datasets for SDGs 11, 15, 16 and 17.}\label{fig:Fig4}
	\end{center}
\end{figure}

Figures \ref{fig:Fig3} and \ref{fig:Fig4} present boxplots illustrating the accuracy obtained through bootstrap sampling for the neural networks trained on the original and augmented datasets. To facilitate visualization, the SDGs were divided into two groups: specifically, goals 3, 4, 8, 9, and 10 (Figure \ref{fig:Fig3}), and goals 11, 15, 16, and 17 (Figure \ref{fig:Fig4}).

The boxplots analysis provide supporting evidence for the choice of the augmented dataset over the original one for neural network adjustment. The observed performance difference is significant, highlighting the effectiveness of data augmentation through clustering, especially in scenarios where the absence of labels is accompanied by a substantial amount of unlabeled data of a similar nature. This aligns well with the context of SDG labels in the Brazilian Supreme Federal Court, as these labels were introduced in 2020, and a substantial volume of texts exists prior to this time point.

It is important to highlight that the synthetic labels responsible for the enhanced performance in the neural networks were not generated to formally expand the originally labeled dataset or replace human assessments. Synthetic labels were applied to originally unlabeled texts for the purpose of investigating improvements in the neural networks – the current classification algorithm in use within the tribunal itself. The classification through clustering does not label all cases within the unlabeled dataset, but rather focuses on cases closer to the centroids of clusters. Therefore, it is not intended to replace more sophisticated classification methods but rather to serve as a complementary tool to enhance overall performance.

\section{Discussion}\label{sec6}

The present work proposes a data augmentation strategy based on text clustering and label propagation, with immediate application in legal contexts. The main idea is to enhance text classification pipelines using augmented datasets through clustering of initially unlabeled documents, i.e., not directly classified by experts. Despite the computational cost, this approach has the potential to replace rapid annotation strategies based on tools like Doccano and Brat (Browser-Based Rapid Annotation Tool). These tools involve human analysis, which can lead to work overlap for employees who work in sectors related to case filing and classification, such as in the Brazilian judiciary sector. The concept serves to allocate experts solely to future classifications, that is, to pieces or cases that will arrive from a certain point onward, while classification algorithms improve their performance with the help of previously unofficially labeled texts.

Using a data set labeled with the 2030 Agenda SDGs by experts from the Supreme Federal Court (STF) and no labels legal text excerpts, it was possible to establish a satisfactory strategy for augmenting the originally labeled dataset, aiming to enhance legal process classification workflows. This activity is already a part of the court's daily operations, supported by machine learning and deep learning through the RAFA 2030 project. This key outcome is accompanied by significant intermediate results, including the increase in positive examples for specific SDGs and the successful use of the doc2vec embedding method in Portuguese-written legal texts (judicial decisions and opinions), for example.

In this work, simple clustering (k-means) and classification (LSTM networks) models were used to establish the foundational baseline for the proposed flowchart presented in Figure \ref{fig:Fig1}. A simple approach was also chosen to propagate synthetic labels, based solely on the proportion of original labels within the established neighborhood. However, other models can be considered at each step of the pipeline. For instance, in future works, BERT models trained on legal contexts in the Brazilian Portuguese \citep{chalkidis2020} and the GPT family \citep{openai2023} could be used to enhance the text abstraction capability in the embedding stage. Adjustments can also be made to other clustering models, such as those from the DBSCAN \citep{dbscan1996}, WEclustering \citep{vivek2021}, and BERTopic \citep{bert2022} families, with the latter two also being based on transformers. Label propagation can be aided by graphical models \citep{koller2009} and by handling legal metadata by legal experts. Finally, numerous classification models can be fine-tuned, and the vast range of methods includes simple approaches based on textual similarity using cosine similarity, ensemble models, and various modern neural network architectures.

\section{Conclusion}\label{sec7}

In this study, data augmentation experiments were conducted to enhance the performance of algorithms tailored for the classification of legal texts according to the Sustainable Development Goals (SDGs) of the 2030 Agenda. The classification involves determining the relevance of specific legal texts to the SDGs themselves. The results demonstrate that some SDGs have increased from fewer than 100 positive examples to around 500 labels, considering both original and synthetic labels in Table \ref{tab4}. This is highly valuable for class balancing and the fine-tuning of machine learning models. Furthermore, improvements in SDG classification performance were observed, with metrics such as accuracy and sensitivity increasing by up to 17\% for SDG 15, for example. The trained doc2vec models also served as an embedding engine for other initiatives within the scope of the Brazilian judiciary.

Further research should explore more sophisticated approaches to embedding, clustering, and classification of legal texts, which may require collaborations with multidisciplinary teams consisting of linguists, legal experts, statisticians, computer scientists, and other professionals. To ensure that impactful legal questions are adequately addressed with the support of machine learning and artificial intelligence, it is always essential to include technical discussions with legal practitioners, whether from courts, academia, or law firms.

\bibliography{sn-bibliography}

\end{document}